\documentclass[preprint,12pt]{elsarticle}

\usepackage{amssymb}

\usepackage{lscape}
 
\usepackage{algorithm}

\usepackage{times} 
\usepackage{helvet}  
\usepackage{courier}  
\usepackage{algorithm}
\usepackage{times}
\usepackage{helvet}
\usepackage{courier}
\usepackage{amssymb}
\usepackage{amsmath}
\usepackage{amsmath, algorithm, algpseudocode}
\usepackage{enumitem}
\usepackage{listings}
\usepackage{xcolor}
\usepackage{booktabs}
\usepackage{multirow}
\usepackage{caption} 
 \usepackage{listings}
\usepackage{xcolor}
\usepackage{tcolorbox}
\usepackage{tcolorbox}

\usepackage{amsmath}

\begin{document}

\begin{frontmatter}



\title{ Can Artificial Intelligence Support Healthcare and Mental Health Through Early Cyberbullying Detection? The Impact of Emotion-Aware AI on Proactive Online Safety  }

%

\author[1]{Hamed Jelodar\corref{cor1}}
\ead{jelodarh@gmail.com}

\author[1]{Amir Firouzi}
\ead{amir.firouzi@unb.ca}

\author[1]{Yen-Wu Lo}
\ead{yenwu.lo@unb.ca}

\author[2]{Maryam Tanha}
\ead{m.tanha@northeastern.edu}

\author[1]{Sajjad Dadkhah}
\ead{sdadkhah@unb.ca}

\cortext[cor1]{Corresponding author}

\affiliation[1]{organization={Faculty of Computer Science},
                addressline={University of New Brunswick},
                city={Fredericton},
                state={NB},
                country={Canada}}

\affiliation[2]{organization={Khoury College of Computer Sciences},
                addressline={Northeastern University},
                city={Vancouver},
                state={BC},
                country={Canada}}

\begin{abstract}
Healthcare systems, mental health, and public well-being are increasingly affected by cyberbullying and harmful online interactions. This paper presents CareGuard, an early-warning framework designed to support healthcare-driven mental health protection and proactive online safety through the detection of cyberbullying-related content using advanced natural language processing techniques. CareGuard integrates zero-shot semantic labeling with fine-tuned transformer-based models, including BERT, DistilBERT, and RoBERTa, to enable robust and context-aware classification across sensitive cyberbullying categories. To improve efficiency and reduce unnecessary computation in healthcare-oriented monitoring settings, the framework incorporates an emotion-aware filtering mechanism alongside cosine similarity–based semantic screening, allowing the system to focus on semantically relevant and emotionally salient content. Experimental results on benchmark datasets demonstrate that CareGuard effectively balances detection accuracy and computational efficiency, highlighting its potential for scalable deployment in healthcare systems, mental health monitoring, and online safety applications.

\end{abstract}


\begin{highlights}
\item Introduces an AI-based system for early cyberbullying detection.
\item Uses emotion-aware filtering to reduce unnecessary processing.
\item Enables early intervention to support mental health.
\item Improves proactive monitoring of harmful online content.
\end{highlights}

\begin{keyword}



 Artificial Intelligence\sep Healthcare Informatics \sep Cyberbullying\sep Society\sep Online Safety \sep emotion

\end{keyword}

\end{frontmatter}




\section{Introduction}
\label{intro}

According to a report by NBC News, a child in Texas died by suicide during an online game, allegedly due to cyberbullying. A 16-year-old boy from Michigan was identified as the suspect and pleaded guilty to aiding suicide and misdemeanor harassment \citep{Ref33}. Mental health challenges associated with social media and cyberbullying are a significant issue that affects individuals of all ages \citep{Ref1,Ref2}. Cyberbullying includes harmful behaviors such as sending abusive messages, posting negative comments, and excluding individuals from online communities. These actions can lead to serious psychological effects, including anxiety, depression, and social isolation \citep{Ref3,Ref4,Ref5}. Compared to traditional bullying, cyberbullying is more difficult to detect and control due to its anonymous and pervasive nature. Children and adolescents are especially vulnerable, as they are highly active on digital platforms \citep{Ref6,Ref7}. To address these challenges, we propose CareGuard, an AI-driven early warning system that leverages large language models (LLMs) and advanced natural language processing (NLP) techniques. The system is designed to detect and categorize cyberbullying and online violence in real time by analyzing digital communications across multiple platforms.

Developing such a system presents several challenges. Ensuring high detection accuracy while minimizing false positives is essential to maintain user trust. In addition, protecting user privacy and adhering to ethical standards are critical considerations. CareGuard addresses these challenges by integrating advanced algorithms that balance detection performance with privacy preservation. The system is designed to be both effective and ethically responsible, making it a valuable tool for enhancing online safety.

\subsection{Research Motivation}
The motivation behind this research stems from the increasing prevalence and severity of cyberbullying and online violence, which pose significant risks to mental health and well-being. The large volume, diversity, and dynamic nature of online communications make timely detection and intervention challenging. Existing methods often focus on static classification and lack the ability to provide early warnings or support proactive intervention. Therefore, there is a need for more robust and scalable approaches that can effectively detect harmful content and facilitate early response.

\subsection{Research Contributions}

The main contribution of this research is the development of CareGuard, an AI-driven early warning framework for detecting cyberbullying and online violence. The key contributions are as follows:

\begin{itemize}
\item We propose a novel NLP-based framework that integrates zero-shot semantic labeling with fine-tuned transformer models for cyberbullying detection.

\item We introduce an emotion-aware filtering mechanism based on emotion annotation and cosine similarity to reduce irrelevant data processing and improve computational efficiency.

\item We developd a multi-task post-modeling analysis module using NLP techniques and prompt engineering to support interpretability and decision-making.

\item We design CareGuard as an early warning system to support mental health–oriented interventions, extending beyond traditional classification-based approaches.
\end{itemize}

\section{Related Works}
\label{sec:1}

The rapid growth of online social platforms has significantly increased exposure to cyberbullying across different age groups, particularly among teenagers. This exposure has been linked to serious psychological consequences, including depression and suicidal ideation \citep{Ref7,Ref8,Ref9,Ref10,Ref11,Ref12}. The World Health Organization recognizes bullying as a major public health concern due to its long-term educational, physical, and mental health impacts \citep{Ref13}.

 \begin{table*}[htbp]
\centering
\caption{Comparison of existing cyberbullying detection studies and the proposed CareGuard framework.}
\label{tab:related_works_comparison}
\resizebox{\textwidth}{!}{%
\begin{tabular}{p{2.3cm} p{3.2cm} p{4.5cm} p{4.2cm} p{2.3cm}}
\hline
\textbf{Study} & \textbf{Main Focus} & \textbf{Main Limitation} & \textbf{Key Insight} & \textbf{Early Warning / Intervention} \\
\hline

\citep{Ref14}
& Cyberbullying detection
& Requires labeled data and may have limited cross-domain generalization
& Combining ML, DL, and BERT can improve detection performance over traditional baselines
& No \\

\citep{Ref19}
& Detection and classification of cyberbullying
& Computationally demanding and dependent on labeled datasets
& Combining BiLSTM and BERT improves contextual representation and supports multi-category classification
& No \\

\citep{Ref22}
& Cyberbullying classification
& Performance may depend strongly on the dataset and domain
& BiLSTM-based models can achieve strong classification performance for cyberbullying detection
& No \\

\citep{Ref23}
& Cyberbullying during COVID-19 using Twitter
& Primarily focused on a specific social-media context and event
& Combining BERT with CNN and MLP provides strong performance for context-specific cyberbullying detection
& No \\

\citep{Ref20}
& Large-scale analysis of abusive tweets
& Focuses mainly on textual abuse and does not provide proactive intervention
& NLP can support large-scale analysis of abusive-language patterns and temporal trends
& No \\

\citep{Ref21}
& Relationship between cyberbullying and suicidal ideation
& Does not provide an automated cyberbullying early-warning system
& Highlights the importance of considering psychological and contextual consequences of cyberbullying
& No \\

\textbf{CareGuard}
& \textbf{Detection \& proactive prevention}
& \textbf{Requires further validation across diverse platforms and real-world settings}
& \textbf{Extends detection toward proactive early warning and intervention support}
& \textbf{Yes} \\

\hline
\end{tabular}%
}
\end{table*}

\subsection{Traditional and Machine Learning Approaches}
Early studies on cyberbullying detection mainly relied on traditional machine learning methods such as Support Vector Machines (SVM), Naive Bayes, and rule-based approaches. These methods use handcrafted features such as word frequency, n-grams, and lexical patterns. While they are simple and interpretable, they have limited ability to capture complex language patterns, sarcasm, and contextual information in social media text.

\subsection{Deep Learning and Transformer-based Methods}
Recent research has shifted toward deep learning models, including Convolutional Neural Networks (CNNs), Recurrent Neural Networks (RNNs), and transformer-based models. In \citep{Ref14}, the authors propose an ensemble approach combining machine learning and deep learning methods, along with a BERT-based model, achieving higher accuracy compared to SVM baselines. Similarly, \citep{Ref19} uses BiLSTM and BERT to detect and classify cyberbullying into categories such as religion, age, and gender. In \citep{Ref22}, different deep learning models are compared, showing that BiLSTM achieves better performance in terms of accuracy and F1-score. Despite these improvements, such models often require large labeled datasets and may not generalize well across different platforms.

\subsection{Sentiment and Context-aware Approaches}
Some researchers focus on sentiment analysis and contextual understanding to improve cyberbullying detection. In \citep{Ref23}, the authors analyze cyberbullying during the COVID-19 pandemic using Twitter data and apply BERT combined with CNN and MLP models, achieving accuracy between 87.2\% and 92.3\%. In \citep{Ref20}, a large-scale analysis of abusive tweets is conducted using NLP techniques to study trends over time. Additionally, \citep{Ref21} highlights psychological factors, showing that psychotic experiences can increase the impact of cyberbullying on suicidal ideation. Although these approaches improve detection, they often focus primarily on text sentiment and may miss deeper behavioral patterns.

\subsection{Limitations of Existing Methods}
Despite significant progress, existing methods face several limitations. First, many models struggle to detect implicit or context-dependent cyberbullying, such as sarcasm or coded language. Second, models trained on specific datasets may not generalize well across different platforms or domains. Third, most approaches focus only on detection and do not provide early warning or intervention mechanisms. Unlike existing approaches that focus solely on classification, the proposed model, CareGuard, introduces an early warning mechanism to support proactive intervention. These advancements position CareGuard as a more comprehensive and practical solution for real-world cyberbullying detection and prevention.

\subsection{Positioning of CareGuard}
\label{sec:careguard_positioning}

The comparison presented in Table~\ref{tab:related_works_comparison} highlights several important gaps in existing cyberbullying detection research. Traditional machine learning methods provide relatively simple and interpretable solutions but rely heavily on handcrafted features and have limited ability to capture contextual and implicit forms of cyberbullying. Deep learning and transformer-based approaches improve contextual representation and classification performance; however, they commonly depend on large labeled datasets and are often evaluated within specific datasets or social-media platforms. Furthermore, most existing studies primarily focus on detecting or classifying cyberbullying rather than supporting proactive responses.

\begin{itemize}
  \item In contrast, CareGuard is designed as a proactive cyberbullying detection and early-warning framework. Rather than limiting the task to binary or multi-class classification, CareGuard aims to identify potentially harmful interactions and provide an early indication of cyberbullying risk. This enables the system to move beyond retrospective detection toward proactive prevention and intervention support. The proposed framework therefore addresses an important gap in existing research by integrating cyberbullying detection with an early-warning mechanism.

 \item As highlighted in Table~\ref{tab:related_works_comparison}, CareGuard is distinguished from previous approaches by explicitly incorporating early warning and intervention support. This design is intended to make cyberbullying detection more practical for real-world applications, where identifying harmful behavior early can be more valuable than detecting it only after the bullying event has occurred. Nevertheless, further evaluation across diverse platforms, datasets, and real-world scenarios is required to assess the generalizability and effectiveness of the proposed framework.

\end{itemize}

\section{Proposed Model}
\label{sec:proposed_model}

In this section, we present \textit{CareGuard}, a multi-phase AI-driven framework for cyberbullying detection and early warning. The proposed model combines text pre-processing, emotion-aware semantic filtering, transformer-based classification, and large language model (LLM)-based post-analysis within a unified pipeline. The main objective of CareGuard is not only to identify cyberbullying content but also to provide interpretable and actionable outputs that can support timely intervention. The overall architecture of the proposed framework is illustrated in Figure~1.

\begin{landscape}
\begin{figure}[h]
\centering
\includegraphics[width=23cm]{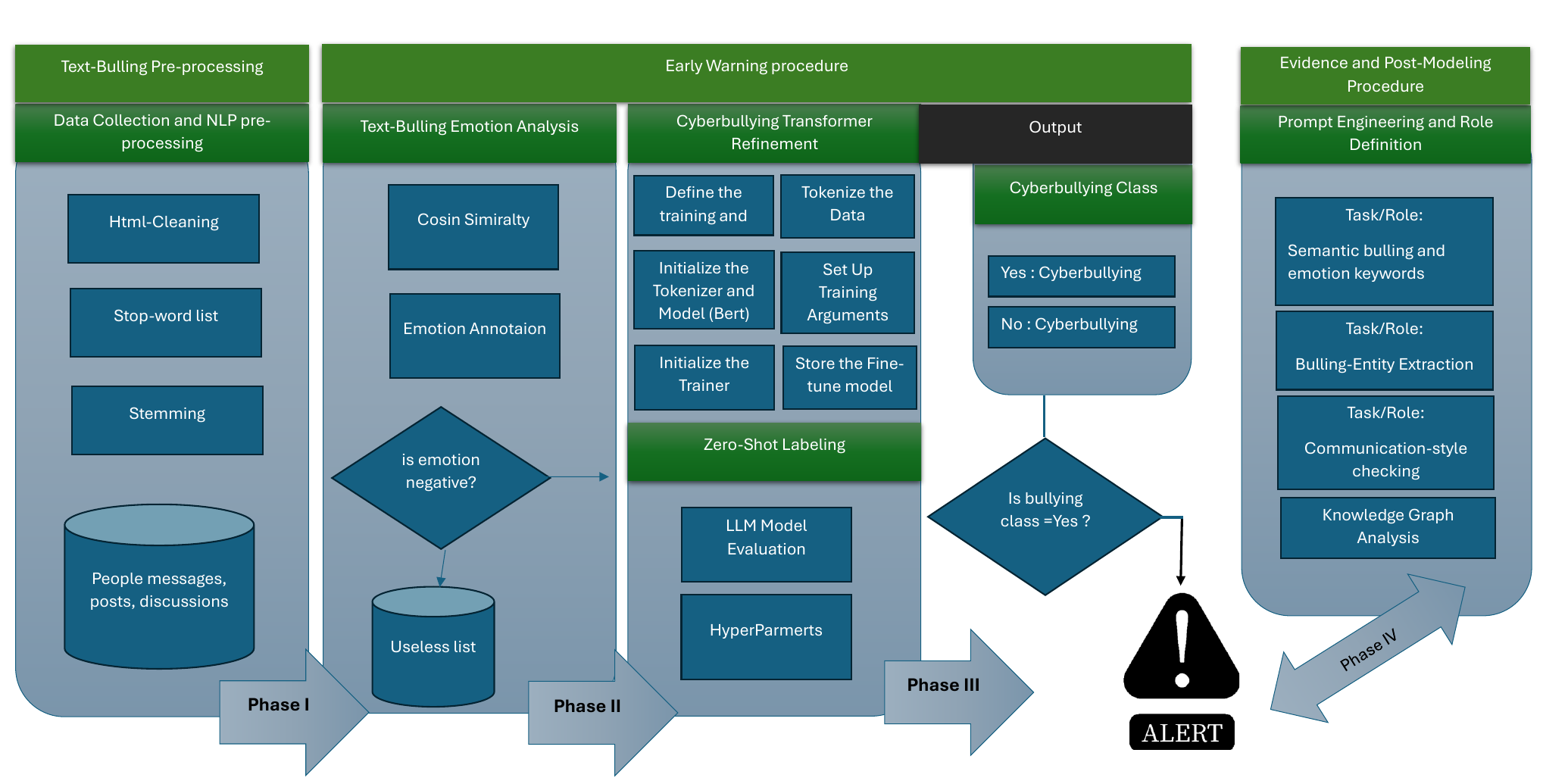}
\caption{Overall architecture of the proposed CareGuard framework, structured into four phases.}
\end{figure}
\end{landscape}

\subsection{Overview of the CareGuard Pipeline}
CareGuard is designed as a four-phase framework: Phase A) Data Collection and Text Pre-processing, Phase B) Semantic Embedding and Emotion-Aware Filtering, Phase C) Transformer-Based Refinement, and Phase D) Post-Modeling LLM Analysis. This design enables CareGuard to operate as an end-to-end early warning system rather than a conventional single-stage classifier.

\subsection{Phase A: Data Collection and Text Pre-processing}
In the first phase, we use an existing dataset \(D\) collected from Kaggle, where \(D = \{T_1, T_2, \ldots, T_N\}\) denotes a set of \(N\) tweets related to cyberbullying. The objective of this phase is to prepare the raw textual data for subsequent semantic and classification analysis.

To improve text quality and reduce noise, several pre-processing operations are applied, including stemming, stop-word removal, and HTML cleaning. These steps standardize the input data and reduce irrelevant variations in the text.

Let \(T_i\) denote the \(i\)-th tweet in the dataset. The stemming operation is defined as:
\[
T_i' = \text{stem}(T_i),
\]
where \(\text{stem}(\cdot)\) reduces each word to its root form. Next, stop-word removal is applied:
\[
T_i'' = T_i' - \text{stop\_words},
\]
where \(\text{stop\_words}\) represents the set of common words with limited semantic contribution. Finally, HTML and noisy markup are removed:
\[
T_i''' = \text{clean\_html}(T_i'').
\]

The processed dataset is then represented as:
\[
D' = \{T_i''' \mid T_i \in D\}.
\]

\begin{algorithm}
\tiny
\caption{Data Collection and Pre-processing}
\begin{algorithmic}
\State \textbf{Input:} Dataset \(D\) with \(N\) cyberbullying-related tweets
\State \textbf{Output:} Processed dataset \(D'\)

\State \textbf{Initialization:}
\State \quad Load dataset \(D\)

\For{each tweet \(T_i \in D\)}
    \State Apply stemming: \(T_i' \gets \text{stem}(T_i)\)
    \State Remove stop-words: \(T_i'' \gets T_i' - \text{stop\_words}\)
    \State Clean HTML/noisy markup: \(T_i''' \gets \text{clean\_html}(T_i'')\)
\EndFor

\State Construct processed dataset:
\[
D' \gets \{T_i''' \mid T_i \in D\}
\]

\State \textbf{Return:} \(D'\)
\end{algorithmic}
\end{algorithm}

\subsection{Phase B: Semantic Embedding and Emotion-aware Filtering}
After pre-processing, the tweets are passed to an emotion-aware semantic filtering module. The goal of this phase is to reduce noise, improve efficiency, and retain tweets that are more likely to contain harmful or relevant content.

For each preprocessed tweet, negative sentiment and emotional cues are extracted. Let \(E(t)\) denote the emotion score of a tweet \(t\) and let \(\theta_{\text{neg}}\) represent the negativity threshold. The filtering operation is formally modeled as a mapping function \(\Phi(t)\) that determines the data flow paths based on emotional intensity:
\[
\Phi(t) =
\begin{cases}
\mathcal{D}_{\text{noise}} \cup \{t\}, & \text{if } E(t) \geq \theta_{\text{neg}} \\
f\big(\mathcal{C}(t), \text{Context}(t), \mathcal{M}_{\text{LLM}}\big), & \text{if } E(t) < \theta_{\text{neg}}
\end{cases}
\]
where \(\mathcal{C}(t)\) denotes the localized structural content of tweet \(t\), \(\text{Context}(t)\) denotes its surrounding contextual or metadata information, and \(\mathcal{M}_{\text{LLM}}\) represents the fine-tuned large language model instance utilized for subsequent, deeper semantic evaluations.

This phase serves two important purposes. First, it removes tweets that are unlikely to contribute meaningful evidence for cyberbullying detection. Second, it prioritizes semantically and emotionally relevant content for deeper analysis in the next stage.

\begin{algorithm}
\tiny
\caption{Emotion-aware Filtering Process}
\begin{algorithmic}
\State \textbf{Input:} Preprocessed tweets
\State \textbf{Output:} Noise dataset \(\mathcal{D}_{\text{noise}}\) and filtered tweets for advanced analysis

\For{each tweet \(t\)}
    \State Extract emotion and sentiment features
    \If{\(E(t) < \theta_{\text{neg}}\)}
        \State Perform advanced semantic analysis using \(f\big(\text{Context}(t), \mathcal{M}_{\text{LLM}}\big)\)
    \Else
        \State Add \(t\) to \(\mathcal{D}_{\text{noise}}\)
    \EndIf
\EndFor
\end{algorithmic}
\end{algorithm}

\subsection{Phase C: Transformer-based Cyberbullying Refinement}
In Phase C, we perform the main cyberbullying classification task using a hybrid transformer-based refinement strategy. The core classifier is a fine-tuned RoBERTa model enhanced with a Bidirectional Gated Recurrent Unit (Bi-GRU) layer. This design combines the contextual representation power of transformers with the sequential modeling capability of recurrent neural networks.

The RoBERTa-based classifier is used as the primary model due to its strong contextual embedding capabilities and robust pre-training strategy. In addition, BERT and DistilBERT are incorporated as supplementary models to further analyze tweets flagged as potentially harmful. This multi-model design improves robustness and provides flexibility under different computational constraints.

\begin{figure}[h]
\centering
\includegraphics[width=14cm]{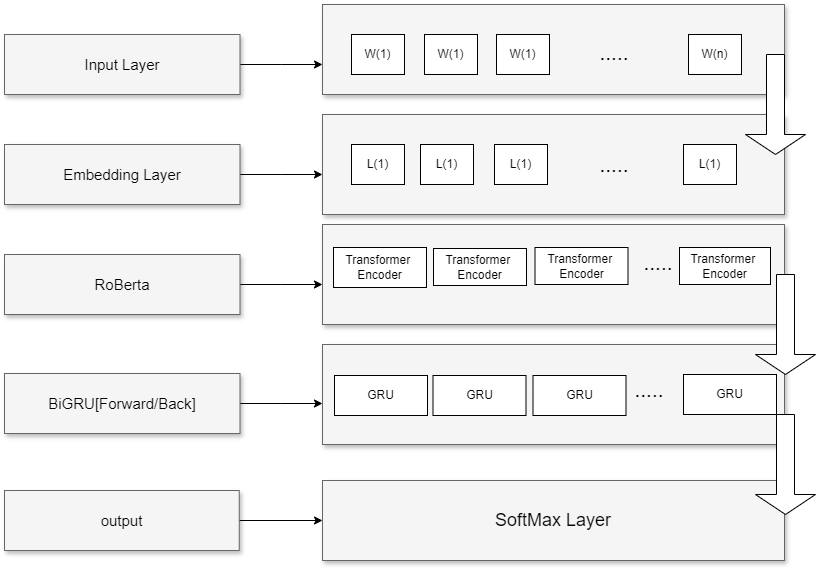}
\caption{Architecture of the transformer-based refinement module using fine-tuned RoBERTa with a Bi-GRU layer.}
\end{figure}

The selected transformer models serve complementary roles:
\begin{itemize}
    \item \textbf{RoBERTa} is used as the primary classifier because of its strong contextual encoding and improved pre-training.
    \item \textbf{BERT} provides effective bidirectional semantic understanding for complex bullying expressions.
    \item \textbf{DistilBERT} offers a lightweight alternative suitable for resource-constrained environments.
\end{itemize}

Let \(M_i\) denote the \(i\)-th transformer model. Its performance is evaluated using accuracy \(A_i\), precision \(Pr_i\), recall \(R_i\), and F1-score \(F1_i\). A weighted performance score is defined as:
\[
S_i = \alpha A_i + \beta Pr_i + \gamma R_i + \delta F1_i,
\]
where \(\alpha\), \(\beta\), \(\gamma\), and \(\delta\) are weighting coefficients.

The aggregated performance across \(n\) models is defined as:
\[
S_{\text{combined}} = \frac{1}{n} \sum_{i=1}^{n} S_i.
\]

Finally, the Bi-GRU layer captures sequential dependencies in the textual input, helping the framework better identify implicit, context-dependent, and semantically connected bullying patterns.

\subsection{Phase D: Post-modeling LLM-based Analysis}
A key contribution of CareGuard is its post-modeling analysis layer. Rather than stopping at a class prediction, the framework further analyzes tweets flagged as cyberbullying using an LLM-based multi-task strategy. This phase improves interpretability and supports practical intervention.

Specifically, we use a LLaMA-based model (\texttt{meta-llama/Llama-2-7b-chat-hf}) to perform three post-modeling tasks:

\begin{enumerate}
    \item \textbf{Mental Health Analysis:} estimates the emotional and psychological impact of the detected bullying language.
    \item \textbf{Bullying-Entity Extraction:} identifies the targeted individuals or groups mentioned in the tweet.
    \item \textbf{Semantic Point Extraction and Summarization:} extracts the most important harmful content and produces a concise explanation.
\end{enumerate}

\subsubsection{Task I: Mental Health Analysis}
This task examines the tone and emotional characteristics of a tweet in order to estimate its potential impact on the victim. Aggression, humiliation, and emotional distress are treated as key indicators of harmful intent.

\subsubsection{Task II: Bullying-Entity Extraction}
This task identifies the entities involved in the harmful interaction, including potential targets of abuse. This enables a better understanding of the social and contextual structure of the bullying event.

\subsubsection{Task III: Semantic Point Extraction and Summarization}
This task extracts the core harmful content and generates a concise summary explaining why the tweet was flagged. This improves interpretability for human analysts and platform moderators.

\begin{figure}[h]
\centering
\includegraphics[width=14cm]{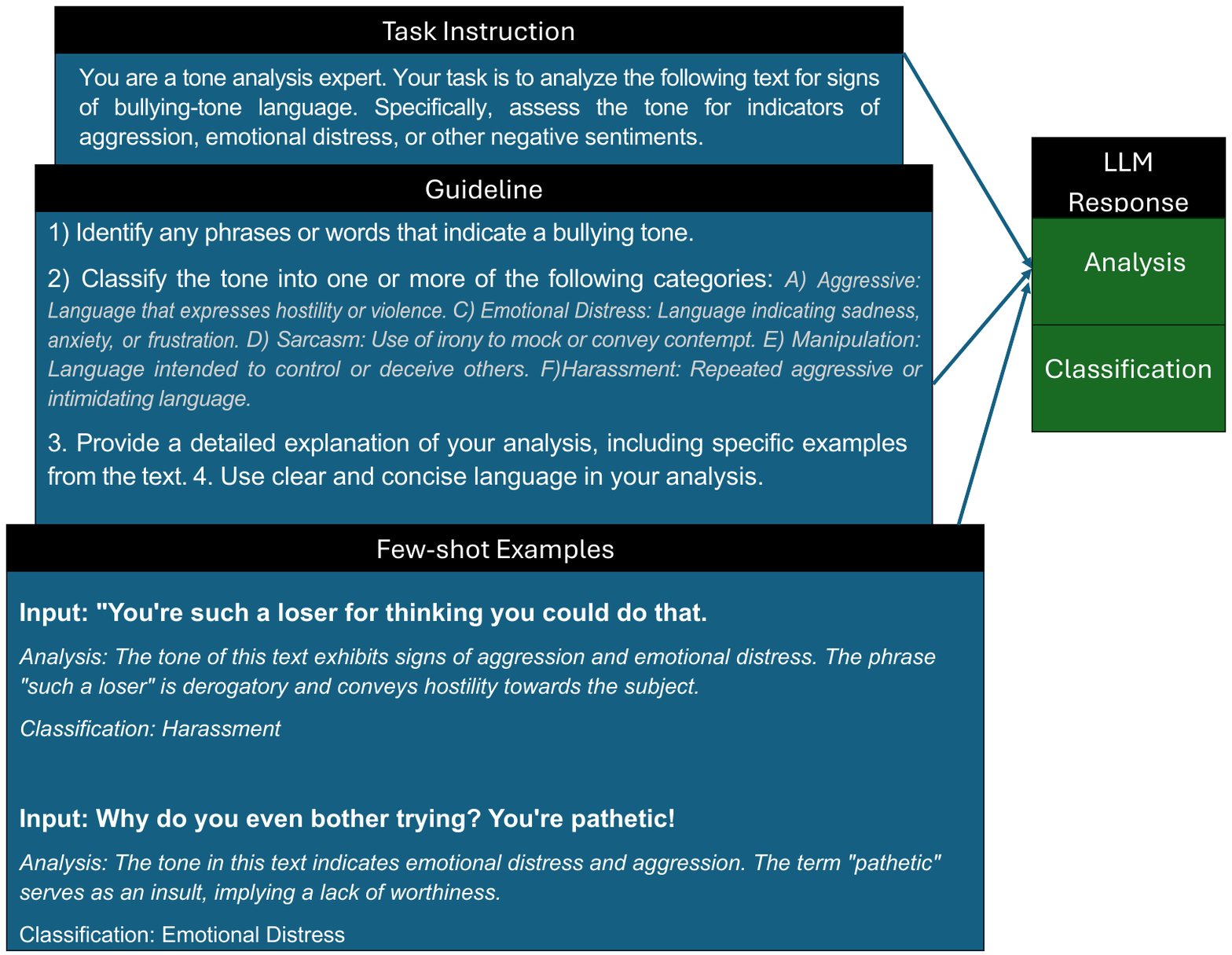}
\caption{Example prompt used in the LLaMA-based post-modeling analysis module.}
\end{figure}

To improve reasoning quality, we incorporate prompt engineering techniques such as Chain-of-Thought (CoT) reasoning and zero-shot/few-shot learning. Chain-of-Thought (CoT) helps the model reason through tone, context, and interaction cues in a step-by-step manner, while zero-shot and few-shot prompting improve generalization to unseen cyberbullying patterns without requiring extensive retraining.

Mathematically, let \(T_{\text{tweet}}\) denote an input tweet classified as cyberbullying. The overall evidence generation process aims to produce robust evidence \(E_{\text{cyber}}\) through three tasks \(T_1, T_2, T_3\):
\[
\tau = f_{\text{tone}}(T_{\text{tweet}})
\]
\[
M_H = f_{\text{impact}}(\tau)
\]
\[
E_{\text{target}} = f_{\text{entity}}(T_{\text{tweet}})
\]
\[
P_{\text{sem}} = f_{\text{sem}}(T_{\text{tweet}})
\]
\[
S_{\text{short}} = f_{\text{summary}}(P_{\text{sem}})
\]

The final output of the post-modeling stage is represented as:
\[
\mathcal{O}_{\text{total}} = \sum_{i=1}^{3} T_i(M_1,\mathcal{C}_{\text{reason}},\mathcal{Z}),
\]
where \(M_1\) denotes the LLaMA model, \(\mathcal{C}_{\text{reason}}\) denotes Chain-of-Thought reasoning, and \(\mathcal{Z}\) denotes zero-/few-shot prompting.

\subsection{Why the Proposed Model is Different}
The proposed CareGuard framework differs from existing cyberbullying detection studies in three major ways. First, it introduces an emotion-aware filtering stage prior to classification, which reduces noise and improves efficiency. Second, it integrates multiple transformer models in a refinement stage rather than relying on a single classifier. Third, it extends beyond conventional detection by incorporating LLM-based post-analysis for interpretability, mental health-oriented assessment, and semantic explanation. As a result, CareGuard functions not only as a classifier but also as an early warning and decision-support system.


\section{Experiment and Settings}

\subsection{Fine-Tuned Candidate Models}

In Phase~D, we implement and evaluate a hybrid cyberbullying detection framework that combines pre-trained transformer models with a Bidirectional Gated Recurrent Unit (BiGRU) layer. The BiGRU component is introduced to enhance the modeling of sequential dependencies in tweets, complementing the contextual representations learned by transformers. We evaluate three widely used transformer architectures—BERT-base, RoBERTa-base, and DistilBERT—each integrated with a BiGRU layer to ensure a fair architectural comparison. All models are fine-tuned using a publicly available cyberbullying dataset from Kaggle \citep{Ref42}. Training is performed with a learning rate of $2 \times 10^{-5}$, a batch size of 32, and early stopping based on validation loss to mitigate overfitting. Model evaluation is conducted on a held-out test set of 1,137 samples. To ensure consistency across models, class labels are aligned by mapping \textit{gender/sexual} to \textit{gender} prior to analysis.\\

\begin{landscape}
\begin{table}[htbp]
\centering
\caption{Comparison among fine-tuned transformer models for cyberbullying detection (RoBERTa-base as reference)}
\label{tab:model_comparison}
\renewcommand{\arraystretch}{1.25}
\setlength{\tabcolsep}{6pt}
\begin{tabular}{lcccccccc}
\hline
\multicolumn{9}{l}{\textbf{Overall Performance (Single-model evaluation)}} \\
\hline
 & \multicolumn{2}{c}{Accuracy} 
 & \multicolumn{2}{c}{Macro F1} 
 & \multicolumn{2}{c}{Weighted F1}
 & \multicolumn{2}{c}{Macro Recall} \\
\cline{2-9}
\textbf{Model} 
 & Estimate & $\Delta$ 
 & Estimate & $\Delta$
 & Estimate & $\Delta$
 & Estimate & $\Delta$ \\
\hline
\textbf{RoBERTa-base} 
 & 0.91 & Ref.
 & 0.90 & Ref.
 & 0.9129 & Ref.
 & 0.88 & Ref. \\

BERT-base 
 & 0.91 & 0.00
 & 0.90 & 0.00
 & 0.91 & $-$0.0029
 & 0.88 & 0.00 \\

DistilBERT 
 & 0.8777 & \textbf{0.0323}
 & 0.8526 & \textbf{0.0474}
 & 0.8748 & \textbf{0.0381}
 & 0.8369 & \textbf{0.0431} \\
\hline
\\[-1.5ex]

\multicolumn{9}{l}{\textbf{Class-level Performance (F1-score)}} \\
\hline
 & \multicolumn{2}{c}{ethnicity/race}
 & \multicolumn{2}{c}{gender}
 & \multicolumn{2}{c}{not\_cyberbullying}
 & \multicolumn{2}{c}{religion} \\
\cline{2-9}
\textbf{Model} 
 & F1 & $\Delta$
 & F1 & $\Delta$
 & F1 & $\Delta$
 & F1 & $\Delta$ \\
\hline
\textbf{RoBERTa-base} 
 & 0.89 & Ref.
 & 0.90 & Ref.
 & 0.95 & Ref.
 & 0.85 & Ref. \\

BERT-base 
 & 0.87 & 0.02
 & 0.88 & 0.02
 & 0.95 & 0.00
 & 0.88 & $-$0.03 \\

DistilBERT 
 & 0.84 & \textbf{0.05}
 & 0.84 & \textbf{0.06}
 & 0.93 & 0.02
 & 0.80 & \textbf{0.05} \\
\hline
\\[-1.5ex]

\multicolumn{9}{l}{\textbf{Error Rate Comparison}} \\
\hline
\textbf{Model} & \multicolumn{2}{c}{Accuracy} & \multicolumn{2}{c}{Error Rate} & \multicolumn{4}{c}{$\Delta$ Error Rate} \\
\cline{2-9}
\textbf{RoBERTa-base} 
 & \multicolumn{2}{c}{0.91}
 & \multicolumn{2}{c}{0.09}
 & \multicolumn{4}{c}{Ref.} \\

BERT-base 
 & \multicolumn{2}{c}{0.91}
 & \multicolumn{2}{c}{0.09}
 & \multicolumn{4}{c}{0.00} \\

DistilBERT 
 & \multicolumn{2}{c}{0.8777}
 & \multicolumn{2}{c}{0.1223}
 & \multicolumn{4}{c}{\textbf{0.0323}} \\
\hline
\end{tabular}

\vspace{1ex}
\footnotesize{
 fine-tuned RoBERTa-based model is used as the reference model.
$\Delta$ denotes the absolute difference relative to the reference.
Positive values indicate worse performance than RoBERTa-base.
All models are evaluated on the same test set of 1,137 samples with aligned class labels.
}
\end{table}

\end{landscape}

\begin{figure}[h]
\centering
\includegraphics[width=9cm]{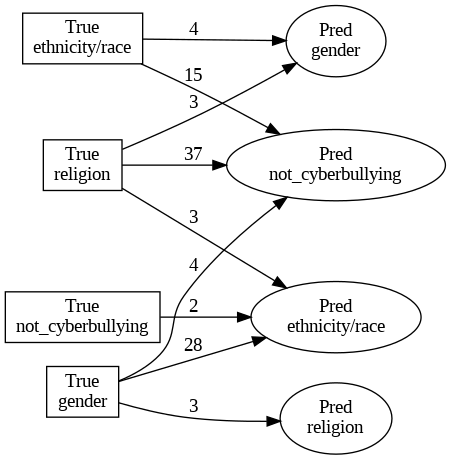}

\caption{Confusion flow visualization of misclassifications for the fine-tuned BERT-based model.}

\label{fig:confusion_matrix}
\end{figure}

\begin{figure}[h]
\centering
\includegraphics[width=9cm]{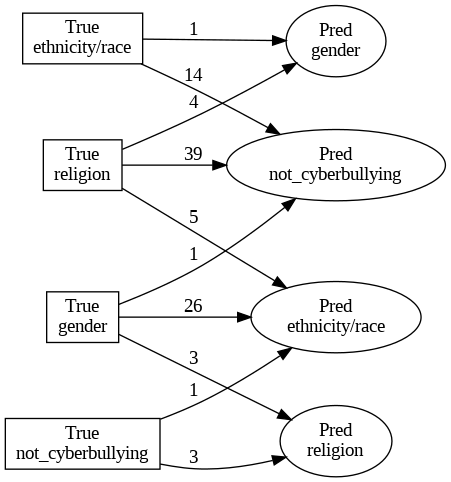}

\caption{Confusion flow visualization of misclassifications for the fine-tuned RoBERTa-based model. }

\label{fig:confusion_matrix}
\end{figure}

\begin{figure}[h]
\centering
\includegraphics[width=9cm]{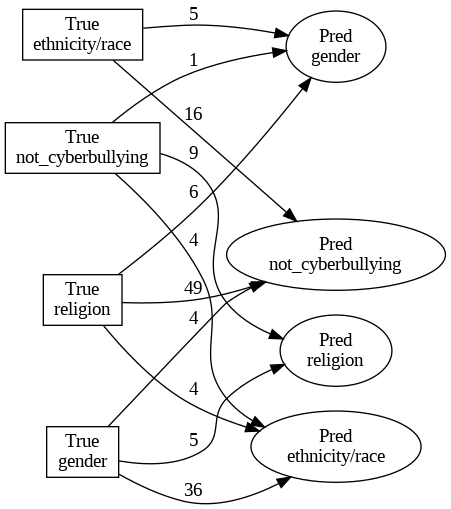}
\caption{Confusion flow visualization of misclassifications for the fine-tuned DistilBERT-based model. }
\label{fig:confusion_matrix}
\end{figure}

\begin{figure}
    \centering
    \includegraphics[width=12cm]{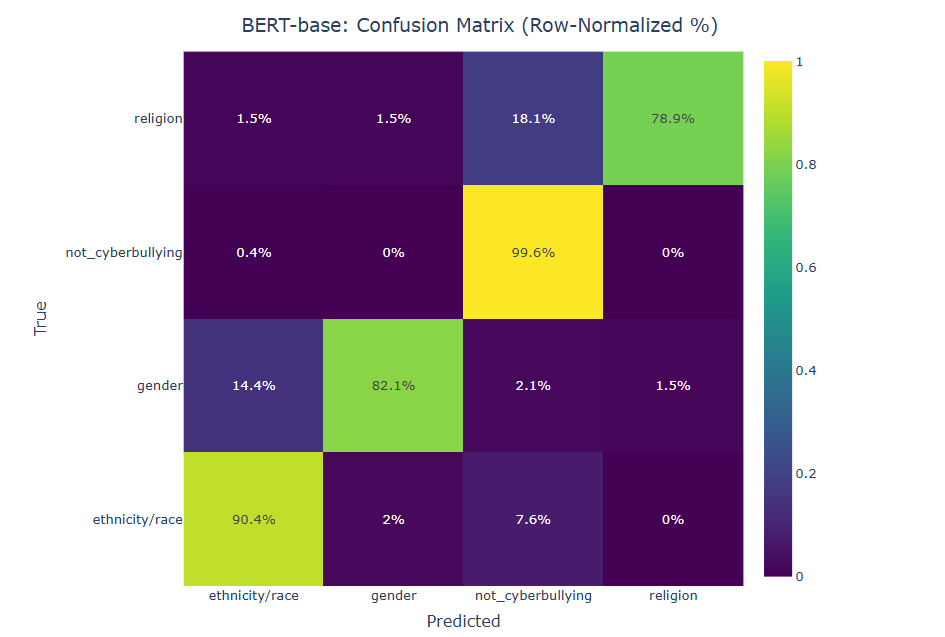}
    \caption{Row-normalized confusion matrix of the fine-tuned BERT-base mode}
    \label{fig:placeholder}
\end{figure}

\begin{figure}
    \centering
   \includegraphics[width=12cm]{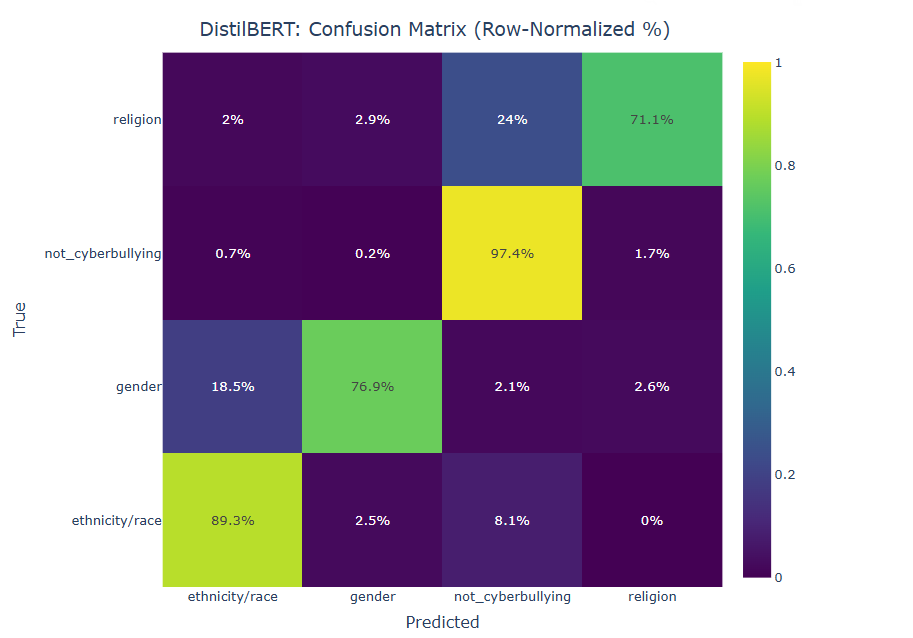}
    \caption{Row-normalized confusion matrix of the fine-tuned DistilBERT model}
    \label{fig:placeholder}
\end{figure}

\begin{figure}
    \centering
    \includegraphics[width=10cm]{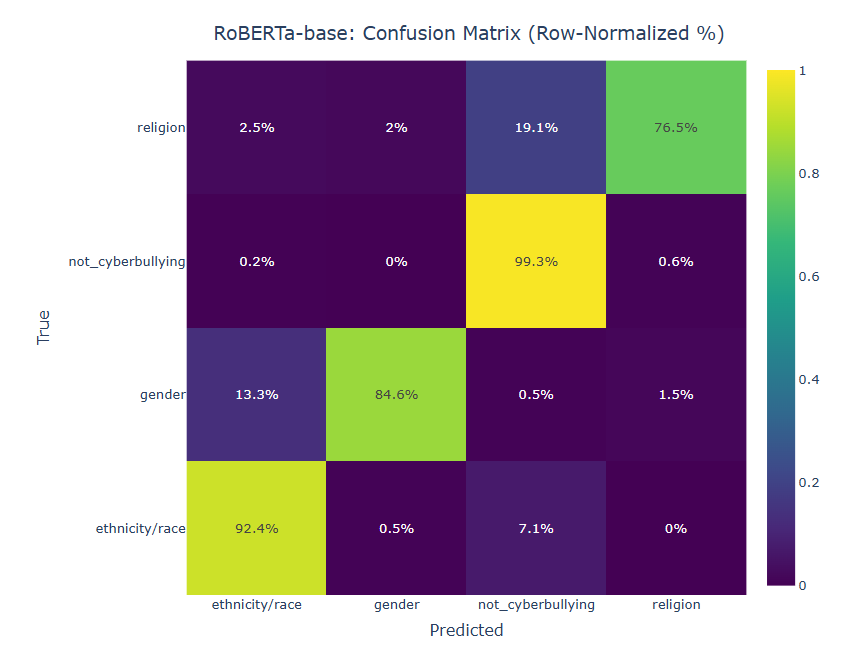}
    \caption{Row-normalized confusion matrix of the fine-tuned RoBERTa-base model.}
    \label{fig:placeholder}
\end{figure}

\vspace{1ex}

\

\subsection{Confusion flow analysis of misclassifications.}
Figures~4, 5, and 6 illustrate the misclassification behavior of the fine-tuned transformer models using complementary visualizations, including top misclassification statistics. Figure~6 presents a confusion flow visualization that highlights only the \emph{misclassification pathways} of the fine-tuned DistilBERT-based cyberbullying detection model. In this diagram, rectangular nodes represent true class labels, while elliptical nodes correspond to predicted labels.

Directed edges connect true and predicted classes, and the numerical annotations on each edge indicate the number of misclassified instances along each path. By excluding correct predictions, the figure provides a focused view of model failures, revealing systematic error patterns rather than overall accuracy. Notably, several minority and sensitive categories (e.g., \textit{religion} and \textit{ethnicity/race}) are frequently misclassified as \textit{not\_cyberbullying}, indicating a tendency of the model to default to the majority class when semantic cues are weak or ambiguous.

\subsection{Performance of Transformer Models}

Table~\ref{tab:model_comparison} provides a comprehensive comparison of the fine-tuned transformer models using RoBERTa-base as the reference. In terms of overall performance, RoBERTa-base and BERT-base achieve identical accuracy (0.91) and Macro F1-score (0.90), indicating comparable global predictive capability. However, RoBERTa-base exhibits a slightly higher weighted F1-score, suggesting improved robustness under class imbalance.

At the class level, RoBERTa-base demonstrates superior performance on sensitive cyberbullying categories, particularly \textit{ethnicity/race} and \textit{gender}, where it achieves the highest F1-scores. These improvements are critical for cyberbullying detection systems, as errors in protected or sensitive categories can have disproportionate social impacts. While BERT-base marginally outperforms RoBERTa-base on the \textit{religion} class, this advantage is offset by its weaker performance on other sensitive categories.

DistilBERT consistently underperforms compared to the other models, exhibiting a notable degradation in Macro F1-score and Macro Recall. The higher error rate observed for DistilBERT indicates a clear trade-off between computational efficiency and detection reliability. This suggests that although DistilBERT may be suitable for resource-constrained environments, it is less appropriate for safety-critical cyberbullying detection tasks.

Overall, RoBERTa-base emerges as the most balanced and robust model, achieving strong global performance while maintaining superior class-level fairness across sensitive categories. Consequently, RoBERTa-base is selected as the primary model for downstream analysis and deployment in the proposed cyberbullying monitoring framework.

\subsubsection{Row-normalized confusion matrices}
Figures~7–9 show the row-normalized confusion matrices for BERT-base, DistilBERT, and RoBERTa-base, all fine-tuned on the cyberbullying dataset. All models achieve high accuracy on the \textit{not\_cyberbullying} class, while gender-based cyberbullying is classified reliably across models. Religion-based cyberbullying remains the most challenging category, often being confused with non-bullying content. RoBERTa-base demonstrates the most balanced performance, particularly for ethnicity/race-based cyberbullying, whereas DistilBERT exhibits higher overall misclassification rates.

\section{Importance of the Application of CareGuard in Real Life}

We believe that the proposed CareGuard model has significant real-world applications in addressing cyberbullying and online violence. Educational institutions such as schools and universities can use this system to monitor online interactions and enhance student safety. Social media platforms, including Facebook and Instagram, can integrate such tools to detect harmful content and enforce community guidelines \citep{Ref39,Ref40}. In addition, technology companies such as Google and Microsoft can incorporate cyberbullying detection features into their products to improve user experience and safety.

Furthermore, non-profit organizations and government agencies can use CareGuard to develop policies and awareness programs aimed at preventing online harassment. Employers can also utilize such systems to detect and mitigate workplace harassment in digital communication channels. In the healthcare sector, hospitals and medical organizations can benefit from cyberbullying detection tools to protect patients, staff, and institutional reputation. These systems enable proactive monitoring of online interactions, helping to maintain a respectful and safe environment for all stakeholders.

\section{Discussion}
\label{sec:discussion}

The experimental results demonstrate that transformer-based models can provide effective performance for cyberbullying detection, while also revealing several challenges associated with sensitive categories, class imbalance, and the practical deployment of automated detection systems. In particular, the comparison among RoBERTa-base, BERT-base, and DistilBERT shows that model architecture and representational capacity influence the balance between classification performance and computational efficiency. The findings also provide insights into how CareGuard can extend conventional cyberbullying classification toward a more proactive early-warning framework.

\subsection{Interpretation of Experimental Results}
\label{sec:interpretation_results}

The experimental results indicate that RoBERTa-base provides the most balanced overall performance among the evaluated transformer models. RoBERTa-base achieved an accuracy of 0.91, a Macro F1-score of 0.90, a weighted F1-score of 0.9129, and a Macro Recall of 0.88. BERT-base achieved comparable overall performance, whereas DistilBERT showed a noticeable reduction across the major evaluation metrics. These results suggest that the stronger contextual representations provided by the full-sized transformer architectures are beneficial for distinguishing between cyberbullying and non-cyberbullying expressions.

The comparable performance of RoBERTa-base and BERT-base is consistent with previous studies that have reported strong performance from transformer-based architectures for cyberbullying detection \citep{Ref14,Ref18,Ref19}. In particular, the ability of transformer models to capture contextual relationships between words is useful for social-media content, where harmful expressions can vary substantially in vocabulary, structure, and context. However, the results also indicate that increasing model efficiency through distillation can involve a measurable performance trade-off. DistilBERT achieved an accuracy of 0.8777 and a Macro F1-score of 0.8526, which were lower than those of both RoBERTa-base and BERT-base.

The class-level results provide additional insight into model behavior. RoBERTa-base achieved F1-scores of 0.89 for ethnicity/race, 0.90 for gender, 0.95 for not cyberbullying, and 0.85 for religion. The relatively lower performance for the religion category suggests that some forms of cyberbullying may be more difficult to distinguish from non-bullying or ambiguous language. Similar challenges have been identified in cyberbullying research, where the linguistic expression of harmful behavior can vary across categories and contexts \citep{Ref1,Ref10,Ref22}. These findings emphasize the importance of evaluating cyberbullying detection systems at the class level rather than relying exclusively on aggregate accuracy.

The confusion-flow analysis further demonstrates that minority and sensitive categories can be more frequently misclassified as not\_cyberbullying. This behavior may be influenced by class imbalance as well as the semantic ambiguity of some harmful expressions. Consequently, a model with high overall accuracy may still produce meaningful errors for specific categories. This observation is particularly important for safety-oriented applications, where errors affecting sensitive categories may have greater practical consequences than errors involving the majority class.

\subsection{Comparison with Previous Studies}
\label{sec:comparison_previous}

The findings of this study are broadly consistent with previous research demonstrating the effectiveness of machine learning, deep learning, and transformer-based methods for cyberbullying detection. Saini et al. \citep{Ref14} demonstrated that combining conventional machine learning, deep learning, and BERT-based approaches can improve cyberbullying detection compared with traditional baselines. Similarly, Gupta et al. \citep{Ref19} investigated BiLSTM and BERT for textual cyberbullying identification and demonstrated the value of contextual representations for multi-category classification. Iwendi et al. \citep{Ref22} also showed that deep learning architectures can provide effective performance for cyberbullying detection.

The results obtained in the present study further support these observations, as RoBERTa-base and BERT-base achieved strong classification performance across the evaluated categories. However, CareGuard differs from conventional classification-oriented approaches by integrating multiple processing stages before and after classification. In particular, the proposed framework combines emotion-aware filtering, transformer-based classification, and LLM-based post-modeling analysis. Therefore, the contribution of CareGuard is not limited to improving classification performance; it is designed to provide additional contextual information that can support the interpretation of detected harmful interactions.

Previous research has also examined cyberbullying within specific social-media contexts and through contextual or psychological perspectives. For example, Sen et al. \citep{Ref23} investigated BERT-based approaches for cyberbullying detection using Twitter data, while Perez and Karmakar \citep{Ref20} examined cyberbullying trends through large-scale analysis of abusive tweets. Fekih-Romdhane et al. \citep{Ref21} further highlighted the relationship between cyberbullying and suicidal ideation. These studies demonstrate that cyberbullying should not be considered solely as a text-classification problem. Instead, its potential psychological and social consequences should also be considered when designing practical detection systems.

In this context, CareGuard extends existing approaches by incorporating an early-warning perspective. Rather than treating the classification result as the final output, the framework performs additional analysis of detected content, including emotional characteristics, targeted entities, and semantic information. This design provides a pathway for transforming a classification result into a more interpretable warning signal that could subsequently be reviewed by human moderators or other responsible decision-makers.

\subsection{Sensitive Categories and Error Analysis}
\label{sec:sensitive_categories}

The class-level evaluation shows that cyberbullying categories are not equally difficult to detect. The relatively strong performance on the not\_cyberbullying class indicates that the models can effectively identify many non-harmful instances. However, the lower performance observed for categories such as religion and ethnicity/race indicates that harmful expressions associated with sensitive topics may be more difficult to classify reliably.\\

One possible explanation is that cyberbullying related to sensitive attributes can be expressed implicitly rather than through explicit abusive terminology. A harmful statement may therefore appear linguistically similar to ordinary discussion when considered without sufficient contextual information. Previous research has similarly emphasized the challenges associated with contextual, implicit, and domain-dependent cyberbullying expressions \citep{Ref1,Ref10,Ref14}. This finding reinforces the importance of contextual modeling and motivates the use of additional semantic and emotion-aware analysis within CareGuard.\\

\subsection{Implications for Early Warning and Mental Health Support}
\label{sec:early_warning_implications}

An important objective of CareGuard is to move cyberbullying detection beyond retrospective classification toward proactive monitoring and early warning. Existing approaches often focus primarily on determining whether a given message belongs to a cyberbullying category \citep{Ref14,Ref19,Ref22}. Although accurate classification is necessary, practical deployment may require additional information regarding the nature, emotional characteristics, and potential target of the harmful interaction. The emotion-aware filtering stage in CareGuard is intended to prioritize content that contains potentially relevant emotional signals before applying more computationally intensive analysis.

This capability is particularly relevant in contexts where cyberbullying may be associated with psychological distress and reduced well-being. Previous research has reported associations between cyberbullying victimization and adverse psychological outcomes \citep{Ref8,Ref12,Ref21}. Therefore, an early-warning system could potentially support timely review and intervention by identifying potentially harmful interactions before they develop into more persistent patterns. However, CareGuard should be considered a decision-support and early-warning framework rather than a clinical diagnostic system. Any real-world healthcare or mental-health application would require additional validation, human oversight, privacy safeguards, and domain-specific evaluation.

\subsection{Practical Implications}
\label{sec:practical_implications}

The findings have implications for the design of automated cyberbullying monitoring systems. First, the strong performance of RoBERTa-base suggests that contextual transformer architectures can serve as effective core classifiers. Second, the lower performance of DistilBERT illustrates the trade-off between computational efficiency and classification reliability. Therefore, the choice of model should depend on the deployment environment and the acceptable balance between latency, computational cost, and detection performance. Third, the class-level results indicate that practical systems should not rely solely on overall accuracy. Finally, the additional analysis provided by CareGuard may help human reviewers understand why content was flagged, potentially supporting more informed moderation and intervention decisions.

\subsection{Limitations}
\label{sec:limitations}

Despite the promising results, several limitations should be acknowledged. First, the framework is trained and evaluated using a single publicly available dataset. Consequently, the observed performance may not fully represent the linguistic diversity, and evolving characteristics of cyberbullying across different social-media platforms. Differences in vocabulary, slang, communication styles, and community norms may introduce domain-shift challenges during real-world deployment. Second, class imbalance remains a challenge, particularly for minority and sensitive categories such as religion and ethnicity/race.

\subsection{Future Work}
\label{sec:future_work}

Future research will extend CareGuard in several directions. First, the framework
will be evaluated across multiple social-media platforms and multilingual
datasets to assess its robustness under domain and language variation.
Cross-domain transfer learning and continual learning will be explored to
reduce the impact of domain drift \citep{Ref43,Ref44,Ref45,Ref46}.

Second, advanced data augmentation, cost-sensitive learning, and debiasing
strategies will be investigated to improve performance on underrepresented
and sensitive categories \citep{Ref47,Ref48}. Fairness-oriented evaluation
will also be conducted to assess consistency across demographic and linguistic
groups.

Third, future versions of CareGuard will incorporate multimodal learning and
conversational context modeling to improve the detection of implicit and
context-dependent cyberbullying. Finally, the complete early-warning pipeline
will be evaluated in realistic scenarios, considering latency, computational
cost, human review, privacy preservation, and responsible intervention
\citep{Ref49,Ref50}. Recent datasets, studies, and state-of-the-art LLM and
GenAI approaches will also be incorporated to ensure the continued relevance
and generalizability of the framework.

\section{Conclusion}
In this study, we introduced CareGuard, an innovative early warning system designed to detect harmful online interactions related to cyberbullying and mental health concerns. The proposed approach integrates emotion-aware filtering, transformer-based classification, and LLM-based post-analysis to improve detection accuracy and interpretability. Experimental results show that RoBERTa-based models achieve the most balanced performance, particularly for sensitive categories. The framework also supports early warning and decision-making, making it suitable for real-world online safety applications. Future work will focus on improving generalization, fairness, and multimodal analysis to enhance robustness across diverse environments.\\\\


\end{document}